%% file: main.tex
\pdfoutput=1
\documentclass[11pt]{article}

\input{add/packages}
\input{add/macros}

\title{The Promises and Pitfalls of LLM Annotations in Dataset Labeling: a Case Study on Media Bias Detection}

\author{
Tomáš Horych\textsuperscript{1}, Christoph Mandl\textsuperscript{1}, Terry Ruas\textsuperscript{1}, André Greiner-Petter\textsuperscript{2}, Bela Gipp\textsuperscript{1}, \\ \textbf{Akiko Aizawa\textsuperscript{2}, Timo Spinde\textsuperscript{1}} \\
\textsuperscript{1}University of Göttingen, Göttingen, Germany \\
\textsuperscript{2}National Institute of Informatics, Tokyo, Japan \\ 
\texttt{t.horych@media-bias-research.org, c.mandl@media-bias-research.org,} \\ 
\texttt{ruas@uni-goettingen.de, greinerpetter@gipplab.org, gipp@uni-goettingen.de} \\ 
\texttt{aizawa@nii.ac.jp, t.spinde@media-bias-research.org}
}
\begin{document}

\maketitle
\begin{abstract}
High annotation costs from hiring or crowdsourcing complicate the creation of large, high-quality datasets needed for training reliable text classifiers. 
Recent research suggests using Large Language Models (LLMs) to automate the annotation process, reducing these costs while maintaining data quality. 
LLMs have shown promising results in annotating downstream tasks like hate speech detection and political framing.
Building on the success in these areas, this study investigates whether LLMs are viable for annotating a complex task of media bias detection and whether a downstream media bias classifier can be trained on such data.
We create \annolexical, the first large-scale dataset for media bias classification with over 48k synthetically annotated examples. Our classifier fine-tuned on it surpasses all of the annotator LLMs by 5-9\% in Mathew’s Correlation Coefficient (MCC) and performs close to or outperforms the model trained on human-labeled data when evaluated on two media bias benchmark datasets (BABE and BASIL).
This study demonstrates how our approach significantly reduces the cost of dataset creation in the media bias domain and, by extension - the development of the classifiers, while our subsequent behavioral stress-testing reveals some of its current limitations and trade-offs.

\end{abstract}

\section{Introduction}\label{sec:intro}

Media bias detection requires high-quality annotations to train classifiers that accurately identify biases across the political spectrum \cite{wessel_introducing_2023, spinde_towards_2021}. Cognitive biases and limited experience often make it hard for raters to annotate bias accurately \cite{spinde_you_2021}, leading to inconsistent annotations across annotators and instances \cite{spinde_neural_2021}. 
Achieving such high-quality annotations is challenging due to the resource-intensive nature of the task and the need for domain expertise \cite{monarch2021human}. 
Popular expert-based datasets in this domain, like BABE \cite{spinde_neural_2021} and BASIL \cite{fan_plain_2019}, contain a limited number of labeled sentences (~4k and ~10k, respectively), with the need for experts and the associated costs limiting their size. 
This limitation, in turn, affects the performance of the resulting models \cite{spinde_neural_2021}. 
The difficulties in creating datasets also affect dataset diversity, which is crucial when improving media bias classification performances \cite{tomas_horych_magpie_2024}. 

\begin{figure*}[ht!]
    \centering
    \includegraphics[width=1\textwidth]{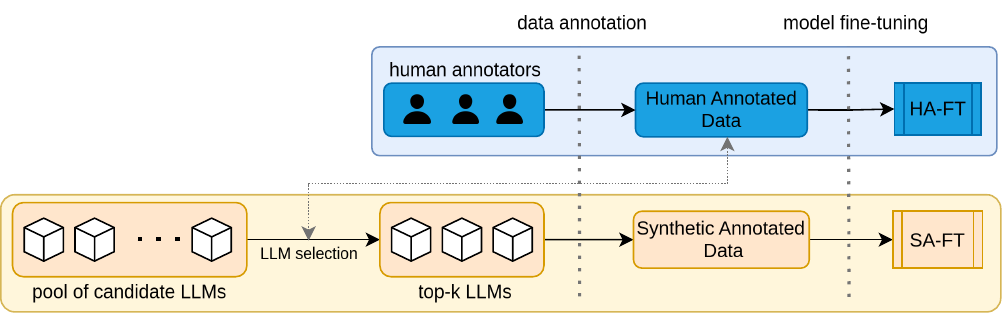}
    \caption{Workflow diagram presenting the difference between the two approaches to fine-tuning the model - \textbf{H}uman-\textbf{A}nnotation {\textbf{F}tine-\textbf{T}uning} (\textbf{HA-FT}) and \textbf{S}ynthetic-\textbf{A}nnotation {\textbf{F}tine-\textbf{T}uning} (\textbf{SA-FT}). The grey arrow between "LLM selection" and "Human Annotated Data" represents an optional step for informed LLM selection.}
    \label{fig:process}
\end{figure*}

Although crowdsourcing is a viable approach to scale data annotation, crowdsource workers often do not have sufficient experience to judge bias correctly \cite{spinde_neural_2021}.
Even more, the quality of crowdsourced labels, particularly from major platforms like Amazon MTurk, has significantly declined over the years \cite{Chmielewski2020}. 
This decline is a common problem in media bias detection and other areas of machine learning and natural language processing (NLP) \cite{Chmielewski2020}.
LLMs offer promising opportunities to support human annotators by automating the annotation process, ensuring consistency, and adapting to specific domains, which can reduce costs and improve or sustain quality. \cite{gilardi_ChatGPT_2023, alizadeh_OpenSource_2023, he_annollm_2023, tan2024largelanguagemodelsdata}.
However, while current research focuses on evaluating LLMs' general capabilities on  NLP benchmarks, the viability of learning from LLM-made annotations in complex downstream tasks like media bias detection remains underexplored. 

In this work, we investigate whether LLMs can provide annotations of sufficient quality to be used to train smaller models for the particular classification task of media bias classification. 
We pick a lexical bias classification as a focal task, as its reliance on the lexical features (see \Cref{sec:focal}) makes it the most popular subtask of a general media bias classification task among media bias researchers in the NLP domain (see \Cref{sec:RW}).
For a detailed overview of media bias and how the components can be defined, we refer to the literature reviews by \citet{RODRIGOGINES2024121641} and \citet{spinde2023media}. 
We introduce a three-stage pipeline to analyze the feasibility of learning lexical bias detection from LLM annotations.
We select three LLMs based on an a priori evaluation, and with a few-shot in-context learning prompt, we label a large-scale training dataset - \annolexical. Finally, we fine-tune a classifier on the aggregated majority-vote label of the \annolexical\ dataset. This approach and its comparison to conventional fine-tuning is depicted in the \Cref{fig:process}.

We compare a Synthetic-Annotations Fine-Tuned classifier (\textbf{SA-FT}) against the conventional classifier fine-tuned on human annotations (\textbf{HA-FT})
in terms of both performance and robustness.
\noindent Our study answers the following research question:
\begin{itemize}
    \item \textbf{RQ1}: Can SA-FT match the performance of the HA-FT on state-of-the-art lexical bias benchmarks?
    \item \textbf{RQ2}: Is the SA-FT classifier robust against spurious correlations?
\end{itemize}

\noindent The contributions of our work are as follows:
\begin{itemize}
    \item We show that the SA-FT classifier outperforms their teacher LLMs and performs comparably with the conventional HA-FT on the sentence-level lexical bias classification task. 
    \item We show that the SA-FT classifier's performance stems from its strength in recalling a major portion of the positive class, but its precision and robustness to input perturbations are worse than that of HA-FT.
    \item We publish the \annolexical\ dataset, a large-scale dataset with 48330 sentences with synthetic lexical bias annotations. 
\end{itemize}

Additionally, we publish a Python package named \annomatic simplifying the annotation pipeline with LLMs, the code, corpus, \annolexical, and the SA-FT classifier publicly available at: \textbf{\href{https://anonymous.4open.science/r/llm-annotations-annomatic-0F23/README.md}{anonymous.4open.science/llm-annotations-annomatic}}\label{s}



\section{Focal task definition}\label{sec:focal}
In this work, we focus on the binary classification of lexical bias at the sentence level.
According to \citet{fan_plain_2019}, lexical bias stems from the choice of words and can be identified based solely on lexical features within a sentence. 
We use a definition of lexical bias instead of linguistic bias as the latter sometimes refers only to morphological text aspects and is generally used with less consistency\cite{spinde2023media}. 
In this work, we will interchangeably use the terms lexical bias and media bias.

\section{Related Work}\label{sec:RW}
\noindent \textbf{Sentence-Level Media Bias Detection.}
Only a few dedicated human-labeled sentence-level media bias detection datasets exist, such as  MBIC ($1700$ sentences) \cite{spinde_mbic_2021}, BASIL ($7919$ sentences) \cite{fan_plain_2019} and BABE ( $4121$ sentences) \cite{spinde_neural_2021}. 
Given the diversity of language and the multitude of options to portray content, especially the small size limits the performance of media bias classifiers, failing to capture its diverse manifestation \cite{wessel-horych-2024-beyond}.
Methods for media bias detection often involve fine-tuning pre-trained language models on these datasets. 
To address the scarcity of ground truth data, researchers have explored various transfer learning strategies, including distant supervision \cite{spinde_neural_2021}, event relation graph augmentation \cite{lei_Sentencelevel_2024}, domain-adaptive pre-training \cite{krieger_domain-adaptive_2022}, fine-grained bias indicators \cite{lin_IndiVec_2024}, and multi-task learning \cite{spinde_exploiting_2022, tomas_horych_magpie_2024}. These transfer learning approaches have consistently yielded positive results, showing the benefits of dataset diversity in the domain.  The top-performing MAGPIE model achieves an F1 score of 0.841 \cite{tomas_horych_magpie_2024} on BABE.
However, these methods address the lack of high-quality training data only indirectly (transfer learning and data augmentation). The main issue—relying on expert-labeled data—remains, as obtaining these labels is both time-consuming and expensive, which limits the necessary scale. This study aims to directly address the problem of sourcing primary data.

\noindent \textbf{LLM dataset labeling.}
Traditional annotation methods face high costs and quality issues \cite{klie_Analyzing_2023, marshall_Who_2023, chmielewski_MTurk_2020}. 
Advances in LLMs suggest they can be efficient alternatives at considerably lower costs. 
Studies show LLMs can match or exceed human annotators in tasks like implicit hate speech detection \cite{ tornberg_ChatGPT4_2023, huang_ChatGPT_2023, he_annollm_2023} and political framing detection \cite{gilardi_ChatGPT_2023, alizadeh_OpenSource_2023}. 
While much research on the topic focuses on ChatGPT (which mostly also shows superior annotation quality across tasks), evaluating open-source models for tasks like media bias detection is crucial to ensure broader accessibility and cost-effectiveness in NLP applications \cite{gilardi_ChatGPT_2023, alizadeh_OpenSource_2023, he_annollm_2023}. 
Across models, experiments demonstrate that few-shot approaches, as well as techniques like Chain-of-Thought (CoT) \cite{wei2022chain} and explanatory methods, significantly improve annotation quality \cite{gilardi_ChatGPT_2023, alizadeh_OpenSource_2023, he_annollm_2023}. 
We discuss in \Cref{sec:discussion} how human input and evaluation will, therefore, still remain crucial for achieving the best results in any approach, including automated annotations.

\section{Annotations with LLMs}
This section describes the process of the synthetic annotation.
Since reproducibility is a significant challenge in the NLP domain \cite{belz-etal-2021-systematic,belz-etal-2023-missing}, we developed \annomatic - a robust tool to make our experiments easily reproducible.
The principal objectives of \annomatic are to 1) abstract away the setup of LLMs from different sources, 2) parse \& interpret the LLM output, and 3) aggregate the results of multiple LLMs in an ensemble.

\subsection{Annotation workflow}
We employ general-purpose LLMs (e.g., LLama2) as annotators, which annotate in a scenario that we refer to as \textit{near-unsupervised}. In this approach, the LLMs generate off-the-shelf annotations with minimal direct human intervention. The only human signal (supervision) provided comes from a set of human-labeled examples included in the prompts, used to guide the model's in-context learning.
We elaborate on the constraint of \textit{near-unsupervision} in \Cref{sec:discussion}. 

The annotation process begins with prompting the annotator LLMs. 
We use a few-shot in-context learning format to prompt the LLM.
The prompt consists of the following components:
\\
\textbf{Examples} - up to 8 examples of human-labeled sentences from a pool of 100 selected sentences. Specifically, we randomly sampled 100 human-labeled examples from the BABE dataset \cite{spinde_neural_2021}, the ground truth dataset for the media bias detection task.
\\
\textbf{Explanations} - alongside each example, an explanation generated by GPT-4 \cite{openai2024gpt4technicalreport} is provided. The explanation is a short text describing how the label in the example was determined.
\\
\textbf{Target of annotation} - the last component is a target sentence to be annotated and a short instruction with label options (e.g., "Contains lexical bias" / "Does not contain lexical bias"). 
\Cref{table:prompt_template} (Appendix) contains the full prompt template used.

The examples and explanations are selected for each sentence instance individually.
In the annotation (inference) time, we retrieve the \textit{k} most similar labeled examples for each target sentence using the KATE algorithm \cite{liu_What_2022} and using the similarity measure as a retrieval criterion, based on the findings of \cite{margatinaActiveLearningPrinciples2023}. Once the LLMs have processed all the data, we parse the responses to extract the final label. 
We search for the most frequently occurring label in the response and match them against a list of \textit{positive} and \textit{negative} label options manually curated by the authors.
If no labels appear in the response or the labels result in a tie, we label it with a question mark '?'.
We later manually review these ambiguous cases and exclude sentences with inconclusive responses.
Finally, we determine the final label for each target sentence via a majority vote among all LLM annotators.

In addition to the open-source code on Github, we release our annotation tool on PyPi\footnote{\url{https://pypi.org/}} under Apache-2.0 license to reduce efforts to replicate our work and simplify its adoption in new projects. 
\annomatic utilizes Haystack\footnote{\url{https://haystack.deepset.ai/}}. This ensures access to state-of-the-art models and easy integration into workflows.

\subsection{Annotator selection}\label{sec:annotator_selection}
To select the LLM annotators, we evaluate a pool of open- and closed-source models on the training set of the BABE dataset. 
The goal of this evaluation is two-fold: to verify that LLMs without fine-tuning can detect lexical bias, thereby qualifying as annotators, and to construct a ranking that we use to select the final annotators.
The candidate LLMs are selected based on the snapshot of the Open LLM leaderboard\footnote{\href{https://huggingface.co/spaces/open-llm-leaderboard/open_llm_leaderboard}{huggingface/open-llm-leaderboard} } at the time of the experiments.
We chose seven open-source general-purpose LLMs from the top of the leaderboard: Falcon 7B Instruct, Zephyr 7B beta, OpenChat 3.5, Mistral-7B-v0.1 Instruct, Mistral-8x7B instruct, LLama 2 7B, 13B and two closed-source models GPT-4-turbo and GPT-3.5-turbo, to cover the closed-source state-of-the-art.
Additionally, we include four models from the FLAN encoder-decoder model family \cite{raffel2020exploring} in sizes ranging from Base to Ultra-Large due to their demonstrated effectiveness in classification tasks \cite{ziems_can_2023}. The list of all models, together with references and basic information, can be found in the Appendix \ref{table:llms}.
The evaluation results are presented in \Cref{table:benchmark}.
While the proprietary GPT-4 outperforms every open-source model, three of the open-source models outperform GPT-3.5 in five prompting settings (marked with $*$).
Due to the cost constraints, we exclude GPT-4, GPT-3.5, and Mixtral-8x7b from the final annotator selection.
We fix the number of selected LLM annotators - $k$ to three - the lowest odd number that will ensure a majority decision while keeping the cost efficiency. Using an odd number of models guarantees a clear majority label. Increasing $k$ to five or more while potentially improving accuracy would result in higher computational costs. 
Therefore, based on the performance, Zephyr 7B Beta, OpenChat 3.5, and LLama 2 13B Chat are selected to annotate the downstream task in the setting with the highest mean performance: 8-shot explanation.
\input{tables/benchmark}

\section{A synthetic bias classifier}

This section presents our proposed process of developing a lexical bias classifier fine-tuned only on the SA-FT.
The process consists of three steps:
1. Curating an annotation corpus, 2. Annotating the corpus, 3. Training a classifier on the synthetic annotations.

\begin{figure}
    \centering
    \includegraphics[width=0.5\textwidth]{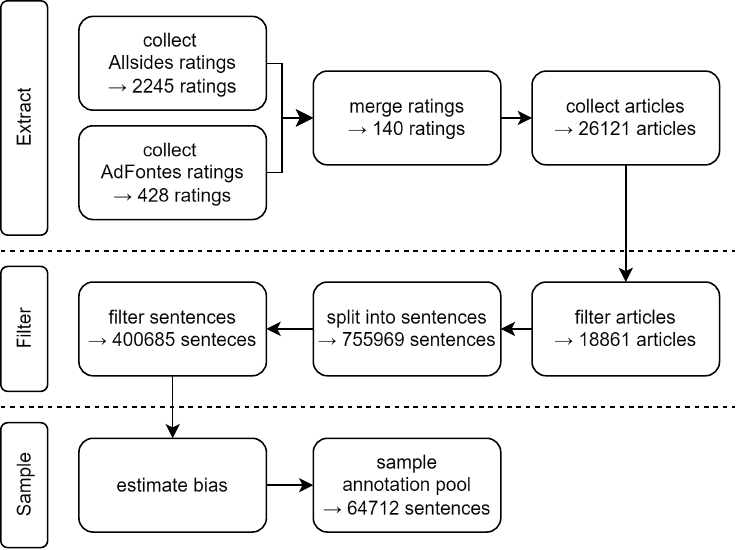}
    \caption{The workflow diagram describing an end-to-end construction of our politically balanced text corpus.}
    \label{fig:epc}
\end{figure}

\subsection{The annotation corpus}\label{sec:annotation_corpus}
This section outlines the process of creating our unlabeled text corpus consisting of news sentences for the downstream annotation.
Given the sensitive nature of bias detection, related work highlights the importance of well-balanced data sources \cite{scheuerman2021datasets, fan_plain_2019}.
An imbalance in the distribution of the political spectrum could lead to skewed models, amplifying existing biases rather than enabling their detection.
We use the platforms \href{https://www.allsides.com/}{allsides.com} and \href{https://adfontesmedia.com/}{adfontesmedia.com} to assess the underlying political leaning of the news text in a left-to-right manner.
\Cref{fig:epc} presents a workflow diagram of the corpus construction.
We break the process down into three parts:
\\

\noindent \textbf{Extract.} We start with scraping all public articles from outlets that have ratings from both \textit{allsides} and \textit{adfontesmedia} platforms. 
Both platforms use left-to-right ratings with different scales; we unify their ratings into five labels: Left, Lean Left, Center, Lean Right, and Right.  
We only keep articles where both platforms agree on the rating.\\
\textbf{Filter.}
We filter out empty, short, or other corrupted articles and keep only articles written in English. 
We then segment these articles into sentences.
Additionally, we trim special characters and other irregularities from the sentences. 
The final collection of filtered sentences contains approximately 400,000 sentences.\\
\textbf{Sample.}
Finally, We sample sentences to ensure the balance across the aforementioned political spectrum. However, we can't ensure a fair distribution of lexical bias before knowing the true labels (i.e., before annotation).
Some outlets may be more likely to contain lexical bias, which could result in an uneven distribution, with one side of the spectrum having mostly biased sentences and the other side being largely neutral after the annotation.

To tackle this issue, we implement a \textit{pre-classification} stage using a state-of-the-art media bias classifier \cite{tomas_horych_magpie_2024} to \textbf{estimate} the sentence's lexical bias before the annotation.

We use this prior bias estimate to sample sentences such that each segment of the political spectrum contains an equal number of sentences, with exactly 50\% estimated to exhibit lexical bias and 50\% exhibiting no bias. This downsampling leads to 64,712 sentences.
This procedure helps to achieve a roughly equal lexical bias distribution across the political spectrum before the costly annotation. 

By estimating the bias and downsampling based on that estimate, we prevent potentially large discarding of sentences after the annotation, given that \citet{tomas2022czechbias} found that, on average, only ~10\% of sentences were biased in a sample of news articles.

\begin{figure*}[h]
    \centering
    \includegraphics[width=1\textwidth]{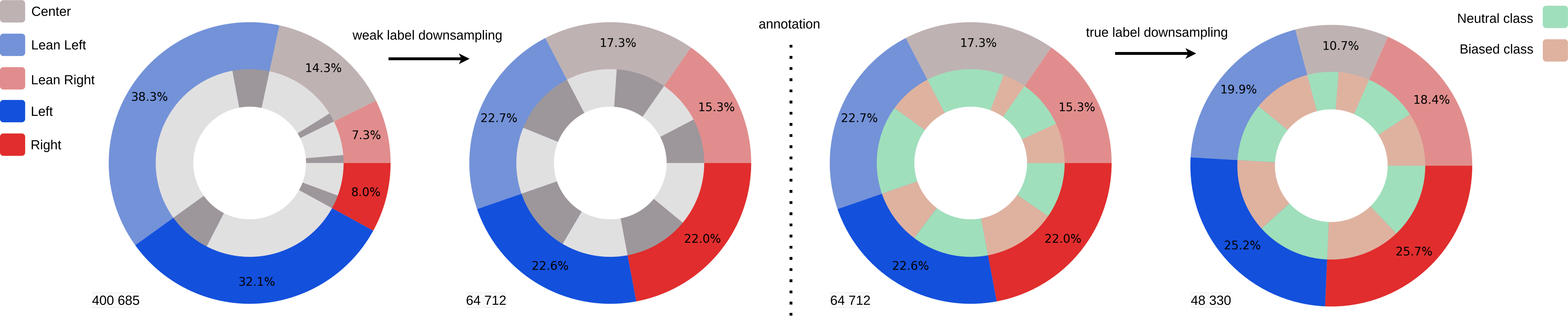}
    \caption{The figure demonstrates the transformation of the unlabeled corpus (left) to the final \annolexical annotated dataset (right) in terms of its size, political ideology distribution, and lexical bias distribution. The inner part of the pie charts represents the distribution of bias labels (neutral/biased) within each part of the political spectrum. The grey depiction of this distribution in the first two plots represents the weak labels estimated before annotation, and the colored (green and red) depiction represents the distribution of the true (annotated) labels.}
    \label{fig:flow}
\end{figure*}

\subsection{Learning from synthetic annotations}
Finally, an ensemble of the three chosen LLM annotators annotates the 64,712 sentences via majority vote (as it is usually done with human annotators).
We use the majority vote instead of exploiting the best-performing model to make our synthetic annotation robust against potential model-specific features and tendencies \cite{navigli2023biases,liang2021towards}. 
This results in 64,712 sentences annotated with lexical bias labels.
We, however, continue to reduce the size of this dataset to ensure an exactly fair distribution of lexical bias labels among the segments of the political spectrum.
For each spectrum segment, we again downsample the sentences, now based on the label obtained through annotation, in a $1:1$ ratio.
The final version of the dataset contains 48,330 sentences.
A diagram summarizing the transformation of the corpus's size and party/label distribution to the final dataset is presented in \Cref{fig:flow}.

We call this final dataset \annolexical, and we make it publicly available on our repository\textsuperscript{\ref{s}}, pre-split into train/dev/test sets with a 0.7, 0.15, and 0.15 proportion, respectively, 

As a last and final step, we fine-tune a RoBERTa\footnote{\href{https://huggingface.co/FacebookAI/roberta-base}{FacebookAI/roberta-base}} encoder LM with a 2-layer classification head on the \annolexical. As our work focuses on comparing two training data scenarios, we keep the model architecture constant to minimize its impact and do not experiment with more models.
We refer to this model as a \textbf{SA-FT} classifier, and we put it to the test in the experiments in the following sections.

\input{tables/performance}

\section{Experiments}
In this section, we present the results of two evaluations of the SA-FT classifier to showcase its properties.
First, we compare the performance of the SA-FT on two well-established lexical bias test sets - BABE and BASIL and compare it to the conventional HA-FT model (\Cref{sec:performance}). 
Secondly, we stress-test the model with a dedicated adversarial test set - CheckList \cite{ribeiro2020beyond}, assessing its robustness against spurious cues and other shortcuts (\Cref{sec:robustness}). 

\subsection{Datasets}\label{sec:datasets}
For the evaluation, we use two key datasets in the sentence-level lexical bias domain: \textbf{BABE} \cite{spinde_neural_2021} - consists of 4121 sentences annotated for binary labels 0 (unbiased) and 1 (biased).
\textbf{BASIL} \cite{fan_plain_2019} - consists of 7919 sentences annotated for ternary labels 0 (unbiased), 1 (lexical biased), 2 (informational-biased). We treat the lexical bias label as a positive class and the informational bias and unbiased as a negative class to unify the task with BABE and \annolexical. 

\subsection{Experimental setup}
For all experiments, we report Matthew's Correlation Coefficient (MCC) as the primary evaluation metric for binary classification due to its higher robustness over the F1 score, as MCC provides a more balanced measure by considering all elements of the confusion matrix \cite{chicco_advantages_2020}. 
For the BABE dataset, we use splits provided by the authors with 75\% of training data and 25\% of test data (1000 sentences). 
We use the entire training set of the BABE dataset to train the HA-FT model and to rank the LLM annotators, as described in \Cref{sec:annotator_selection}. 
We then use the BABE test set for the evaluations. 
From the BASIL dataset, we use all 7919 sentences for the evaluations.
We execute all experiments and annotations on one Nvidia A100 GPU. All training and evaluations were run as a single run.

\subsection{A downstream SA-FT generalization}\label{sec:performance}

In this experiment, we evaluate the generalization ability of the SA-FT classifier in three settings:
\begin{itemize}
    \item \textbf{Comparison with the teacher models.} We evaluate the SA-FT classifier on the BABE test set against the three LLM annotators that annotated its training data.
    \item \textbf{Comparison with the HA-FT.} We compare the BABE test performance of SA-FT against HA-FT.
    \item \textbf{Performance on out-of-distribution test set.} Finally, we compare the SA-FT and the HA-FT classifiers on the held-out BASIL dataset.
\end{itemize}
\input{tables/checklist}

\noindent The evaluation results are presented in \Cref{table:performance}. We report the following findings.
First, we observe an improvement of 2.3\% in the SA-FT performance over the majority vote of the annotators and 5-9\% compared to the single LLMs.
We want to point out that the SA-FT classifier is smaller than the original annotators and was trained on their majority vote.
This result demonstrates that our proposed framework achieves a 5\% improvement over the best of the chosen LLM annotators (LLama 2 13B chat) while reducing the cost of deployment by a factor of 100\footnote{The LLama 2 13B Chat has 13 billion parameters, while our RoBERTa SA-FT classifier has roughly 130 million.} or 300 if a majority vote is used.
We attribute the gap between the majority-voted label and the SA-FT performance to the generalization from the synthetic annotations.

While the SA-FT classifier generalizes from the synthetic annotations, the HA-FT classifier, fine-tuned on the BABE training data, still outperforms the synthetic model by 1,5\%. 
Because the HA-FT model has the advantage of being tested on data from the same distribution as its training set\footnote{Both sets are two different splits of one dataset.}, we also evaluate both models on the held-out BASIL dataset.
In this evaluation, the SA-FT classifier outperforms the HA-FT model by $3.1\%$. We verify this result with the McNemar paired test for labeling disagreements \cite{gillick1989some} and find it statistically significant (with p<0.05). However, we find that both models perform relatively poorly, and the SA-FT model only recalls a slightly larger portion of the positive class.
This low score can be partially explained by the observation that both models often classify the information bias class as positive.

Finally, because \annolexical\ is larger (34k) than the BABE training set (3k), we create a coreset of the \annolexical, with the same size (3k) by following the approach of \cite{chai_Efficient_2023} and fine-tune another model on this coreset. We denote this model as SA-FT\textsubscript{coreset}.
In the fair comparison regarding training size, we observe that the SA-FT falls short and underperforms the HA-FT by 4\% on the BABE test set. However, while it still performs better on the BASIL dataset, its performance is even more skewed to low precision and high recall.

\subsection{Robustness against shortcuts} \label{sec:robustness}
While the SA-FT classifier can match the HA-FT model in raw performance, lexical bias detection is subtle, and the conventional performance metric may only partially capture the model's behavior. 
Therefore, we adapt the idea of \textit{CheckList} - a behavioral stress-testing of the classifiers \cite{ribeiro2020beyond} and extend its prior adoption in the media bias domain \cite{wessel-horych-2024-beyond}.
Inspired by the original \textit{CheckList}, we use three high-level tests: MFT (Minimum Functionality Test), INV (Invariance Test), and DIR (Directional Expectation Test).
Please refer to the work by \citet{ribeiro2020beyond} for further information about the \textit{CheckList} method. 
We then use the \textit{CheckList} to again, compare the SA-FT and HA-FT classifiers. 
The full description of each test case with examples and the results of each model are presented in \Cref{table:checklist}.

The SA-FT demonstrates a minor advantage (1.4\%) in the minimum functionality test and a significant advantage (20+\%) in the directional expectation test, where we expected the introduction of loaded words to change the neutral label to positive.
In other words, the SA-FT is more attentive to strongly connotated words (e.g., \textit{shockingly}, \textit{terrible}). 
We also argue that these results align with generally lower recall of the HA-FT classifier in \Cref{sec:performance}. 
However, the HA-FT prevails on every invariance test, which tests the models' sensitivity to input perturbance. 
These results show that while the SA-FT method achieves results on par with HA-FT in MCC metric, it falls short in robustness to input changes and is less precise than conventional HA-FT.

\section{Discussion}\label{sec:discussion}
In this section, we discuss the role of humans in classifier development.
In this study, we showed that LLMs can effectively annotate datasets for a task as complex as media bias detection and that a downstream classifier achieves comparable results with a model trained on human-labeled data.
However, the proposed framework only relies on a \textit{near-unsupervised} regime. 
While annotations are automated by the LLMs, there are two crucial touchpoints of human interaction in the process. 
First, LLMs are selected based on their ranking on a dedicated human-labeled development set. 
Without this evaluation, practitioners will either have to rely on general NLP benchmarks, which may not reflect a good ranking for their specific task, or resort to random selection.
Second, we prompt LLMs with human-labeled examples to enable in-context learning. 
Although this requires only a small number of annotations, it requires domain expertise and annotation effort.
Lastly, the result of the behavioral testing shows a significant gap between the robustness of models trained on synthetic and human-made annotations.
This indicates a need to improve the model's resilience to subtle changes in input. 
One possible way to tackle this is to augment the synthetic training process with human-made adversarial examples or increase the human effort in de-biasing the underlying dataset before the annotation (e.g., pruning/randomizing the named entities).
While LLMs hold great potential, human intervention is still essential in automated annotation, especially for tasks such as media bias, both in the role of a guide and evaluator.

\section{Conclusion}\label{sec:conclusion}
In this paper, we investigated the viability of using Large Language Models as annotators for training datasets to tackle the need for more high-quality resources in the media bias classification domain.
We showed that general-purpose LLMs can generate reasonable annotations off-the-shelf, and we used three LLM annotators to create the first large-scale dataset for lexical bias classification - \annolexical - with 48330 sentences.
We subsequently show that a classifier fine-tuned on the \annolexical synthetic annotations can match and even outperform a conventional model trained on human annotations while reducing the cost and effort required for human annotations.
While our new model performs competitively on two media bias benchmarks, it falls short in classification precision and robustness against input perturbations.
This defect becomes especially apparent when we scale down the size of \annolexical to match the size of the existing gold-standard dataset.

In our future work, we aim to evaluate the scaling laws of the synthetic annotations and the role of diversity in the underlying dataset.
We hypothesize that the number of synthetic annotations can be exploited further, possibly leading to better and more robust models with the potential to transfer our results to other classification problems.

\section*{Limitations}\label{sec:limitations}
As our approach to first annotate and then classify lexical bias relies directly on using state-of-the-art LLMs, one limitation is the computational cost of running the very large models.
We acknowledge that our limited computational resources prevented us from testing and utilizing the most advanced models (those with more than 50 billion parameters and proprietary models).
These cutting-edge models require immense computational power for inference but could potentially enhance performance.
Secondly, we only evaluate the whole pipeline with three LLM annotators selected greedily based on the benchmark. We did not evaluate other combinations of the annotators due to the computational restrictions. However, since we evaluate the downstream model robustness and out-of-distribution generalization, another run with a random selection of LLMs would bring more insight into how the selection affects the downstream classifier behavior.

\section*{Ethics Statement}\label{sec:ethics}
Media bias strongly depends on personal perception, making it a sensitive issue, especially in the context of automated annotations.
Some bias forms depend on factors other than the content, e.g., a different text perception due to a reader’s background.
While in this paper, we merely investigate the possibilities of automated data annotation if used within a publicly available classifier, quality control of what is classified as bias, especially when subjective, is a main part of our ongoing and future work. 
We recognize the potential for introducing bias in model training and annotation processes and have attempted to mitigate these through diverse data sources and balanced representation.
We see no immediate risk to our work; however, we note that current models still make false predictions and discourage potential users from using them in production. 
By automating the annotation process, we aim to make the dataset creation in the media bias domain less expensive, which, together with additional quality control, will ideally lead to larger availability of media bias classifiers. 
We also believe that creating dedicated datasets and classifiers for individual tasks will result in lower energy consumption than running resource-expensive LLMs locally.

Lastly, we want to declare that the authors used ChatGPT during the writing process of this work, primarily for minor rephrasing and grammar correction.

\section*{Acknowledgements}
This work was supported by the Lower Saxony Ministry of Science and Culture and the VW Foundation.
Furthermore, this project was funded by the Deutsche Forschungsgemeinschaft (DFG, German Research Foundation) – 554559555 and some of the results partially by EXIST-Gründungsstipendium. Finally, the authors would like to express their gratitude towards Prof. Dr. Michael Granitzer for consulting and valuable advices.

\bibliography{anthology,custom}

\appendix

\input{tables/llms}

\section{Prompting}
\label{sec:appendix_prompt}
\begin{table}[h!]
\begin{tabular}{p{\linewidth}}
You are an expert in media bias. \\
\{ for TEXT, LABEL, EXPLANATION in examples \}\\
Instruction: '[TEXT]'\\
Classify the sentence above as BIASED or NOT BIASED.\\
Output: Let's think step by step. [EXPLANATION] The answer is [LABEL].\\
\{ endfor \}\\
Instruction: '[SENTENCE]'\\
Classify the sentence above as BIASED or NOT BIASED.\\
Output: Let's think step by step.  
\end{tabular}
\caption{Prompting Template in pseudo-code. \{..\} indicates a command.}
\label{table:prompt_template}
\end{table}

\section{Data Contamination}
We tested the contamination of GPT-4-Turbo and GPT-3.5 Turbo by following \cite{golchin2024} on a sample of the BABE dataset. Our tests indicated that no contamination is present for both models.

\end{document}

%% file: add/packages.tex
\usepackage[]{acl}
\usepackage{times}
\usepackage{latexsym}

\usepackage[T1]{fontenc}

\usepackage[utf8]{inputenc}

\usepackage{microtype}

\usepackage{inconsolata}

\usepackage{graphicx}
\usepackage{times}
\usepackage{latexsym}
\usepackage[T1]{fontenc}
\usepackage[utf8]{inputenc}
\usepackage{microtype}
\usepackage{inconsolata}
\usepackage{graphicx}
\usepackage{adjustbox}
\usepackage{tabularx}
\usepackage{arydshln}
\usepackage{comment}
\usepackage{courier}
\usepackage{xcolor}
\usepackage{amssymb}
\usepackage{hyperref}
\usepackage{float}
\usepackage{array}
\usepackage{multirow}
\usepackage{makecell}
\usepackage{xcolor}
\usepackage{listings}
\usepackage{cleveref}
\usepackage{svg}

%% file: add/macros.tex
\definecolor{todocolor}{HTML}{FF5733}

\newcommand{\annolexical}{\textit{Anno-lexical }}
\newcommand{\annomatic}{\textit{Annomatic }}

\definecolor{lightgray}{rgb}{0.9, 0.9, 0.9}

%% file: tables/benchmark.tex
\begin{table*}[h!]
\centering
\begin{adjustbox}{width=1\textwidth}
\begin{tabular}{l|ccccccccc|c}
\hline
model & 0-shot & + sys prompt & 0-shot Exp & 2-shot & 4-shot & 8-shot & 2-shot Exp & 4-shot Exp & 8-shot Exp & mean \\
\hline
Zephyr 7B beta & \textbf{0.551}* & 0.385 & 0.369 & 0.538 & 0.548 & 0.558 & 0.6 & 0.616 & 0.627 & 0.532\\
OpenChat 3.5 & 0.389 & 0.499 & \textbf{0.503} & 0.577 & 0.581 & \textbf{0.593}* & 0.565 & 0.58 & 0.622 & \textbf{0.546}\\
Mistral-7B-v0.1 Instruct & 0.343 & 0.357 & 0.248 & 0.353 & 0.415 & 0.46 & 0.487 & 0.495 & 0.534 & 0.41\\
LLama 2 7B Chat & 0.15 & 0.101 & 0.294 & 0.359 & 0.416 & 0.497 & 0.554 & 0.581 & 0.579 & 0.392\\
LLama 2 13B Chat & 0.238 & 0.032 & 0.325 & 0.406 & 0.448 & 0.517 & 0.619 & 0.619 & 0.613 & 0.424\\
Flan-UL2 & 0.489 & \textbf{0.534} & 0.462 & 0.532 & 0.526 & 0.537 & 0.432 & 0.459 & 0.516 & 0.499\\
Falcon-7B-Instruct  & 0.052 & 0.038 & 0.128 & 0.175 & 0.227 & 0.178 & 0.344 & 0.304 & 0.274 & 0.191\\
FLAN-T5-XL  & 0.302 & 0.356 & 0.346 & 0.406 & 0.415 & -  & -  & -  & -  & 0.365\\
FLAN-T5-Large  & 0.133 & 0.312 & 0.335 & 0.165 & 0.146 & -  & -  & -  & -  & 0.218\\
FLAN-T5-Base & 0.107 & 0.12 & 0.061 & 0.044 & 0.044 & -  & -  & -  & -  & 0.075\\
\hdashline
Mixtral-8x7B Instruct & 0.277 & 0.279 & 0.494 & \textbf{0.583} & \textbf{0.595}* & 0.588 & \textbf{0.646}* & \textbf{0.654}* & \textbf{0.662} & 0.531\\
GPT-3.5 Turbo & 0.511 & 0.596 & 0.56 & 0.595 & 0.586 & 0.591 & 0.624 & 0.633 & 0.663 & 0.595\\
GPT-4 Turbo & 0.683 & 0.697 & -  & 0.71 & 0.699 & 0.7 & 0.83 & 0.786 & 0.753 & 0.732\\
\hline
average & 0.325 & 0.331 & 0.344 & 0.419 & 0.434 & 0.522 & 0.57 & 0.572 & \textbf{0.584} &\\ 
\hline
\end{tabular}
\end{adjustbox}
\caption{All results are measured with Mathew's Correlation Coefficient (MCC) on the BABE train/development (combined) set. Bold scores mark the best-performing open-source model for a given prompting. An asterisk * marks performance higher than GPT-3.5. The blank spots ($-$) mark runs where A) the size of the model's context window is insufficient and B) the model's output diverges from the instruction.}
\label{table:benchmark}
\end{table*}

%% file: tables/performance.tex
\begin{table*}[ht]
\begin{adjustbox}{width=1\textwidth}
\begin{tabular}{l|c|cccc|cccc}
\hline
 & & \multicolumn{4}{c|}{\textbf{BABE\textsubscript{test}}} & \multicolumn{4}{c}{\textbf{BASIL}} \\
\hline
 & size\textsubscript{train} & P & R & F1 & MCC & P & R & F1 & MCC  \\
\hline
Zephyr 7B beta & 8-shot & 0.831 & 0.773 & 0.801 & 0.569 & - & - & - & - \\
OpenChat 3.5 & 8-shot & 0.814 & 0.825 & 0.819 & 0.588 & - & - & - & - \\
LLama 2 13B Chat & 8-shot & 0.828 & 0.834 & 0.831 & 0.614 & - & - & - & - \\
majority vote & - & 0.852 & 0.823 & 0.837 & 0.639 & - & - & - & - \\
\hdashline
SA-FT  & 34k & 0.875 & 0.814 & 0.843 & 0.662 & \textbf{0.171} & 0.502 & \textbf{0.254} & \textbf{0.205} \\
HA-FT & 3k & \textbf{0.916} & 0.772 & 0.838 & \textbf{0.678} & 0.169 & 0.384 & 0.235 & 0.174 \\
\hdashline
SA-FT\textsubscript{coreset} & 3k & 0.829 & \textbf{0.859} & \textbf{0.844} & 0.638 & 0.136 & \textbf{0.696} & 0.228 & 0.201 \\
\hline
\end{tabular}
\end{adjustbox}
\caption{The results of the evaluation of the LLM annotators, SA-FT classifier, and HA-FT on two lexical bias benchmark test sets. The highest values within each column are marked in bold.}
\label{table:performance}
\end{table*}

%% file: tables/checklist.tex
\begin{table*}
\centering
\begin{adjustbox}{width=1\textwidth}
\begin{tabular}{|c|c|l|c|c|}
\hline
\textbf{Test Type} & \textbf{Test} & \textbf{Examples} & \textbf{HA-FT} & \textbf{SA-FT} \\
\hline
MFT & \makecell[l]{\textbf{1. factual test}: Short neutral sentences from \\ fact-checking datasets should be unbiased.} & "What is a stereotype? An unfair, generalization about a group of people." & 0.951 & \textbf{0.964}  \\
\hdashline
INV & \makecell[l]{\textbf{1. locations test}: Replace locations \\ with random locations should not change the label.}  & \makecell[l]{"\colorbox{lightgray}{Hawaii} $\rightarrow$ \colorbox{lightgray}{U.S.} eyes even stricter gun laws in wake of shooting that \\  killed 2 police officers."} & \textbf{0.984} & 0.971\\
 & \makecell[l]{\textbf{2. pronouns test}: Replacing named entities \\ with He/She/Them should not change the label.} & \makecell[l]{Despite \colorbox{lightgray}{Portman's} $\rightarrow$ \colorbox{lightgray}{her} insistence that she has tried to advance female \\ directors, only one of her feature films was directed by a female.} & \textbf{0.971} & 0.957 \\ 
 & \makecell[l]{\textbf{3. prejudice test}: Replacing one minority \\ with other minority should not change the label.} & "For some people, \colorbox{lightgray}{Buddha} $\rightarrow$ \colorbox{lightgray}{Christ} holds immense significance." & \textbf{0.895} & 0.852 \\
\hdashline
DIR & \makecell[l]{\textbf{1. loaded-words test}: Injecting biased adjectives \\ and biased adverbs should change neutral to biased.} & \makecell[l]{"The EU has \colorbox{lightgray}{shockingly} secured up to \colorbox{lightgray}{outrageous} 400 million doses\\ of AstraZeneca ’s experimental vaccine."} & 0.413 & \textbf{0.664}\\
\hline
\end{tabular}
\end{adjustbox}
\caption{This table shows examples and results of the \textit{CheckList} stress-testing of the two models. Model trained on human labels - \textbf{HA-FT} and the one trained on synthetic labels - \textbf{SA-FT}. The examples for each test represent instances where the model with the lower score on the right failed and the other succeeded. 
The formatting and style of this table are inspired by the tables used in the original \textit{CheckList} paper \cite{ribeiro2020beyond}.}
\label{table:checklist}
\end{table*}

%% file: tables/llms.tex
\begin{table*}
\centering
\begin{adjustbox}{width=1\textwidth}
\begin{tabular}{l|ccccccccc|c}
\hline
name & source & availability & parameters & link \\
\hline
Zephyr 7B beta & \citet{tunstall2023zephyr} & open & 7B & \href{https://huggingface.co/HuggingFaceH4/zephyr-7b-beta}{HuggingFaceH4/zephyr-7b-beta} \\
OpenChat 3.5 & \citet{wang2023openchat} & open & 7B & \href{https://huggingface.co/openchat/openchat_3.5}{openchat/openchat-3.5}\\  

Mistral-7B-v0.1 Instruct & \citet{jiang2023mistral} & open & 7B & \href{https://huggingface.co/mistralai/Mixtral-8x7B-Instruct-v0.1}{mistralai/Mixtral-8x7B-Instruct-v0.1}\\  

LLama 2 7B chat & \citet{touvron2023llama2openfoundation} & open & 7B & \href{https://huggingface.co/meta-llama/Llama-2-7b-chat}{meta-llama/Llama-2-7b-chat} \\ 
LLama 2 13B chat & \citet{touvron2023llama2openfoundation} & open & 13B & \href{https://huggingface.co/meta-llama/Llama-2-13b-chat}{meta-llama/Llama-2-13b-chat} \\  
Falcon-7B-Instruct & \citet{falcon40b} & open & 7B & \href{https://huggingface.co/tiiuae/falcon-7b-instruct}{tiiuae/falcon-7b-instruct} \\
Flan-T5-base & \citet{flan} & open & 248M & \href{https://huggingface.co/google/flan-t5-base}{google/flan-t5-base} \\
Flan-T5-large  & \citet{flan} & open & 783M & \href{https://huggingface.co/google/flan-t5-large}{google/flan-t5-large} \\
Flan-T5-XL& \citet{flan} & open & 2.8B & \href{https://huggingface.co/google/flan-t5-xl}{google/flan-t5-xl}\\
Flan-UL2 & \citet{yitayew2023flan} & open & 20B & \href{https://huggingface.co/google/ul2}{google/ul2} \\
Mixtral-8x7B Instruct & \citet{mistral2024mixtral} & open & 13B (MoE inference) & \href{https://huggingface.co/mistralai/Mixtral-8x7B-Instruct-v0.1}{Mixtral-8x7B-Instruct-v0.1}\\  
GPT-3.5 Turbo & \citet{openai2023gpt35} & closed & unknown & \href{https://platform.openai.com/docs/models/gpt-3-5-turbo}{openai/gpt-3-5-turbo}\\
GPT-4 Turbo & \citet{openai2024gpt4technicalreport} & closed & unknown & \href{https://platform.openai.com/docs/models/gpt-4-turbo}{openai/gpt-4-turbo} \\  
\hline
\end{tabular}
\end{adjustbox}
\caption{All LLMs evaluated listed with meta data.}
\label{table:llms}
\end{table*}